\documentclass[10pt,twocolumn,letterpaper]{article}

\usepackage{cvpr}              

\usepackage{graphicx}
\usepackage{amsmath}
\usepackage{amssymb}
\usepackage{booktabs}
\usepackage{enumitem}
\usepackage{xcolor}
\usepackage{makecell}

\usepackage{multirow}

\usepackage[pagebackref,breaklinks,colorlinks]{hyperref}

\usepackage[capitalize]{cleveref}
\crefname{section}{Sec.}{Secs.}
\Crefname{section}{Section}{Sections}
\Crefname{table}{Table}{Tables}
\crefname{table}{Tab.}{Tabs.}

\def\cvprPaperID{4057} 
\def\confName{CVPR}
\def\confYear{2022}

\begin{document}

\title{Align and Prompt: Video-and-Language Pre-training with Entity Prompts}

\author{Dongxu Li$^{1,2}$,~~
Junnan Li$^1$,~~
Hongdong Li$^2$,~~
Juan Carlos Niebles$^1$,~~
Steven C.H. Hoi$^1$ \\$^1$Salesforce Research, $^2$The Australian National University\\
{\tt\small dongxuli1005@gmail.com},~{\tt\small \{junnan.li,jniebles,shoi\}@salesforce.com},~{\tt\small hongdong.li@anu.edu.au}
}


\maketitle

\def\DX#1{{\color{red}{\bf [DX:}{\it{#1}}{\bf ]}}}
\def\HL#1{{\color{blue}{\bf [HLi:}{\it{#1}}{\bf ]}}}

\def\name{\textsc{AlPro}}

\def\Vec#1{{\boldsymbol{#1}}}
\def\Mat#1{{\boldsymbol{#1}}}
\begin{abstract}
Video-and-language pre-training has shown promising improvements on various downstream tasks. Most previous methods capture cross-modal interactions with a standard transformer-based multimodal encoder, not fully addressing the misalignment between unimodal video and text features. Besides, learning fine-grained visual-language alignment usually requires off-the-shelf object detectors to provide object information, which is bottlenecked by the detector's limited vocabulary and expensive computation cost. 

In this paper, we propose Align and Prompt: a new video-and-language pre-training framework (\name), which operates on sparsely-sampled video frames and achieves more effective cross-modal alignment without explicit object detectors. First, we introduce a video-text contrastive (VTC) loss to align unimodal video-text features at the instance level, which eases the modeling of cross-modal interactions.
Then, we propose a novel visually-grounded pre-training task, prompting entity modeling (PEM), which learns fine-grained alignment between visual region and text entity via an entity prompter module in a self-supervised way. Finally, we pretrain the video-and-language transformer models on large webly-source video-text pairs using the proposed VTC and PEM losses as well as two standard losses of masked language modeling (MLM) and video-text matching (VTM). 
The resulting pre-trained model achieves state-of-the-art performance on both text-video retrieval and videoQA,
outperforming prior work by a substantial margin.
Implementation and pre-trained models are available at \small\url{https://github.com/salesforce/ALPRO}.

\end{abstract}
\vspace{-1em}
\section{Introduction}
\label{sec:intro}
Video-and-language pre-training aims to jointly learn multimodal representations that transfer effectively to downstream tasks, such as {\em text-video retrieval} and {\em videoQA-video question answering}.
Compared with images, videos usually contain more redundancy in consecutive frames. This challenges models on both capacity and computation efficiency.
Most prior approaches~\cite{li2020hero,miech2019howto100m,zhu2020actbert,miech2020end,sun2019videobert,luo2020univl} circumvent the expensive computation overhead by using offline-extracted video features. Since the video feature extractors are fixed without finetuning, these approaches are suboptimal when transferring to distinct target domains.
In contrast, recent emerging approaches~\cite{lei2021less,Bain21} sample frames sparsely from videos,
which enable end-to-end pre-training and finetuning of video backbones.
In this work, we adopt the sparse video-text pre-training paradigm considering their effectiveness on downstream tasks.

%


\begin{figure}
    \centering 
\includegraphics[width=0.48\textwidth]{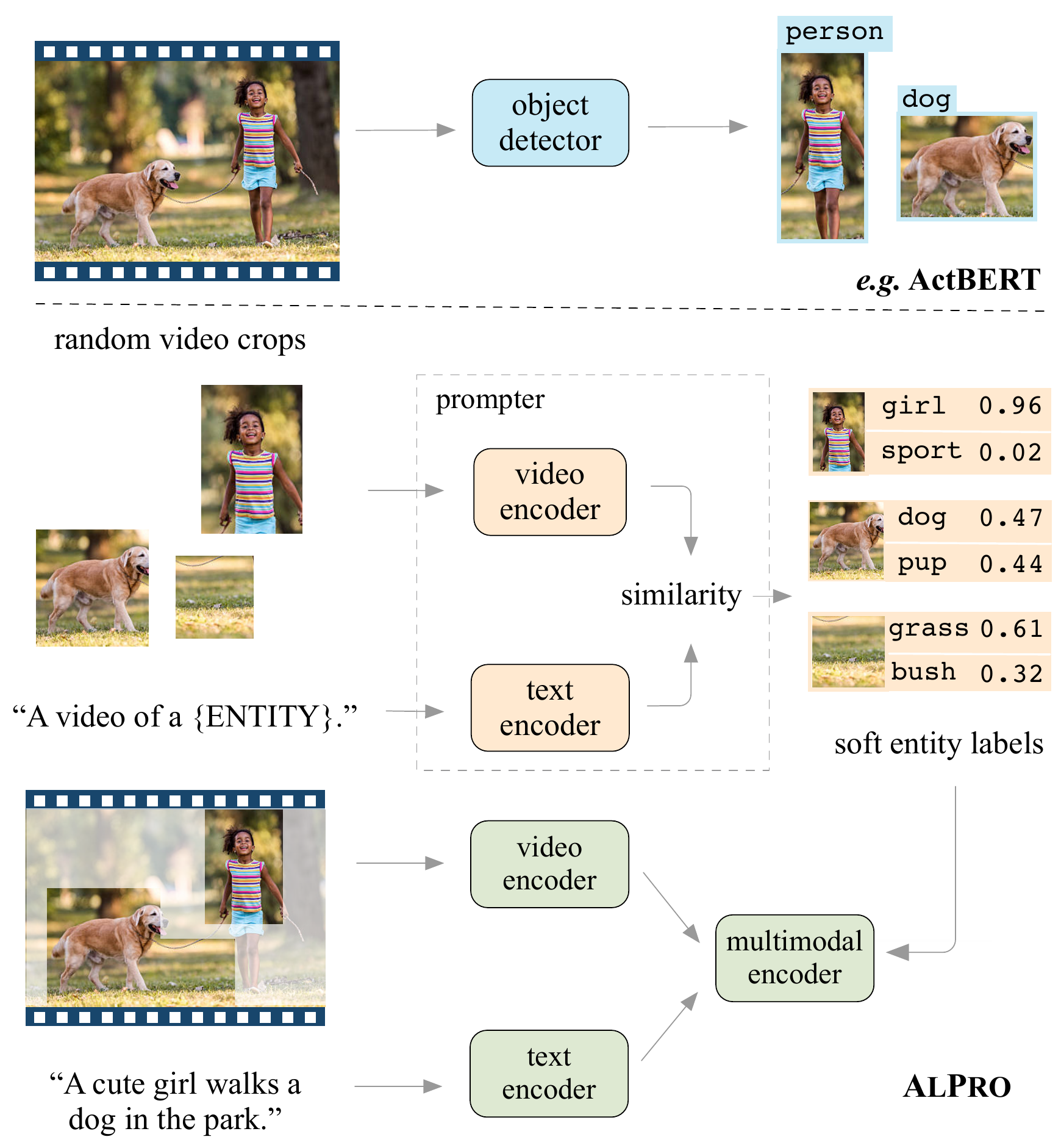}
    \vspace{-15pt}
    \caption{Generating supervision for region-entity alignment. \textbf{Above}: previous methods (\eg~ActBERT\cite{zhu2020actbert}) rely on object detectors with expensive computation cost and limited object categories, leaving text data unexploited. \textbf{Below}: \name~generates soft entity labels with a prompter module, which computes similarities between video crops and textual entity prompts. \name~requires no detector while taking advantage of video-text alignment to generate entity labels with a large vocabulary, thus strengthening the cross-modal learning.}
    \label{fig:teaser}
    \vspace{-1em}
\end{figure}

Despite their promising performance, current video-text pre-training models have several limitations.
%
(1) The interaction between video and text features is commonly modeled trivially using either dot-product~\cite{miech2019howto100m,miech2020end,Bain21,xu2021videoclip} or cross-modal transformer encoders~\cite{sun2019videobert,zhu2020actbert,li2020hero,lei2021less}. 
However, features from individual modalities typically reside in different embedding spaces.
Such misalignment makes it less effective to directly model cross-modal interaction.
(2) Many visually-grounded pre-training tasks~\cite{sun2019videobert,li2020hero} do not explicitly model fine-grained regional visual information (e.g. objects),
which proves important for downstream tasks emphasizing on visual reasoning (e.g. videoQA).
Although there are attempts which employ object detectors~\cite{zhu2020actbert,uniter} to generate pseudo-labels as supervision, they suffer from imprecise detections and a restricted number of object categories. For example, detectors trained on MSCOCO~\cite{lin2014microsoft} recognize less than a hundred different categories.
%
(3) The previous sparse pre-training model~\cite{lei2021less} is trained with image-text pairs using an image encoder, which makes it less effective in modeling temporal information. 

In this paper, we tackle these challenges with a new video-and-language pre-training framework: Align and Prompt (\name). 
Architecture-wise, \name~first encodes frames and text independently using a transformer-based video encoder and a text encoder,
and then employs a multimodal encoder to capture cross-modal interaction.
\name~learns both instance-level video-text alignment and fine-grained region-entity alignment.
The instance-level alignment is learned by applying a video-text contrastive loss (VTC) on the unimodal features,
which encourages paired video-text instances to have similar representations.

%

In order to better capture fine-grained visual information and strengthen region-entity alignment, \name~introduces a new visually-grounded pre-training task, called \emph{prompting entity modeling},
where we ask the video-text model to predict entities appearing in randomly-selected video crops using jointly video and text inputs (see Figure~\ref{fig:teaser}).
To address the unavailability of entity annotations, we design a standalone \emph{entity prompter} module that generates reliable pseudo-labels.
%
Specifically, the entity prompter consists of two unimodal encoders to extract video and text features, respectively.
We first train the entity prompter using only VTC loss and freeze its parameters thereafter.
Then during pre-training, we feed video crops and text prompts (\eg ``\texttt{A video of \{Entity\}.}'') to the prompter, where each \texttt{Entity} is from the frequent nouns appearing in the pre-training corpus.
We then compute the normalized similarity between the entity prompts and the video crop as the pseudo-label to supervise the pre-training.
%

\textbf{Our key contributions are}:
\textbf{(1)}
	We introduce \name, a video-language pre-training method that is the first to learn effective cross-modal representations from \emph{sparse} video frames and texts. 
\textbf{(2)}	We introduce a video-text contrastive loss to better align instance-level unimodal representations, thus easing the modeling of cross-modal interaction.
\textbf{(3)}	We propose a novel visually-grounded pre-training task, prompting entity modeling, that enables the model to capture fine-grained region-entity alignment.
\textbf{(4)}
	We demonstrate the effectiveness of \name~on both video-text retrieval and videoQA. \name~significantly improves over previous state-of-the-art methods, for example, achieving $3.0\%$ and $5.4\%$ absolute lift in recall scores on the finetuning and zero-shot text-video retrieval task on MSRVTT. 


\begin{figure*}[ht]
\centering
  \includegraphics[width=\textwidth]{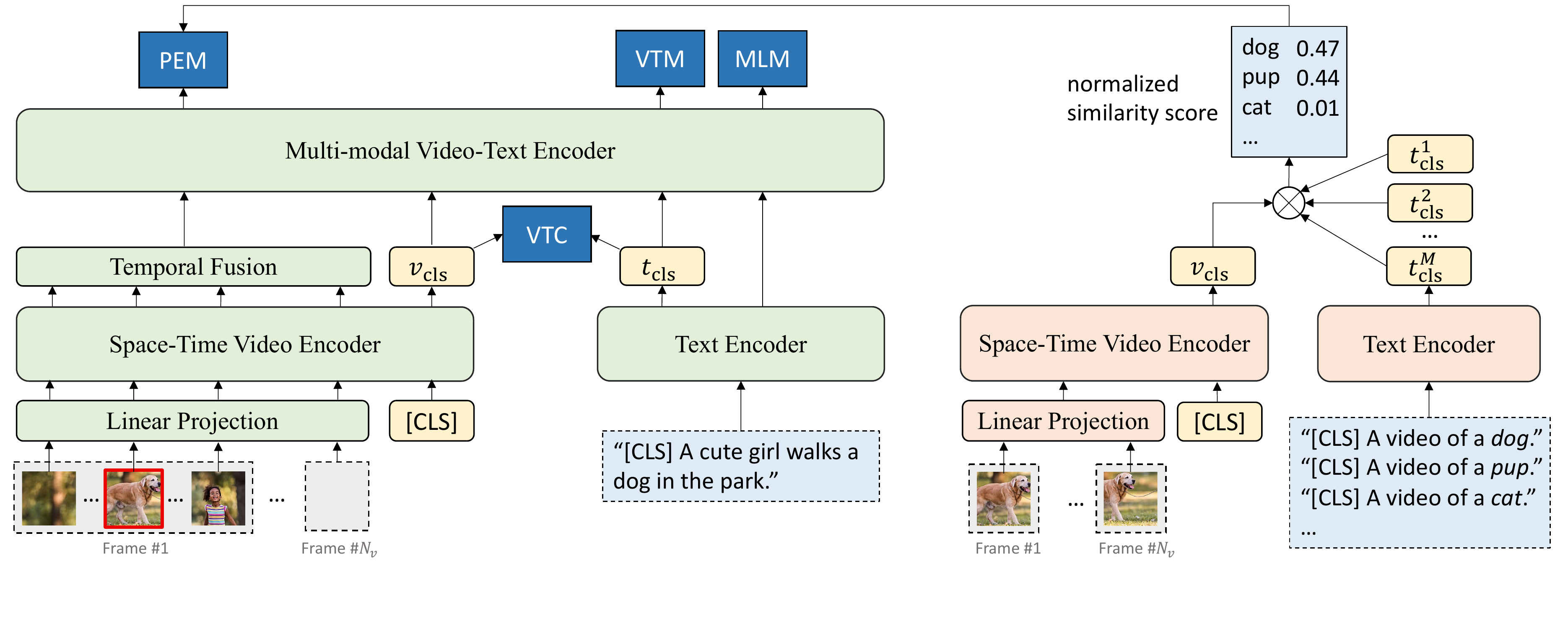}
\vspace{-7ex}
\caption{\name~pre-training framework. \textbf{Left}: the \textit{video-language pre-training model} contains a space-time video encoder, a text encoder, and a multi-modal encoder, all of which are transformer-based. Besides two canonical objectives masked language modeling (MLM) and video-text matching (VTM), we introduce video-text contrastive loss (VTC) to learn instance-level video-text alignment, and prompting entity modeling (PEM) to learn fine-grained region-entity alignment. 
\textbf{Right}: the \textit{prompter} which generates soft entity labels as supervision for PEM. The prompter consists of frozen unimodal encoders that are trained with VTC. During pre-training, it produces similarity scores between a randomly-selected video crop and a set of text prompts instantiated with entity names.
}
\vspace{-1em}
\label{fig:arch}
\end{figure*}

\section{Related Work}\label{sec:relatedwork}
\noindent\textbf{Dense \emph{versus} Sparse Video Representation.}~Consecutive frames in videos usually contain visually similar information. 
Such redundancy opens up a research question on how to learn effective video-and-language representations without excessive computation overhead.
Most prior methods on text-video retrieval~\cite{xu2016msr,yu2018joint,liuuse,gabeur2020multi,patrick2021supportset} and videoQA~\cite{gao2018motion,lei2018tvqa,fan2019heterogeneous,le2020hierarchical} employ pre-trained visual backbones and extract video features for each frame \emph{densely} yet offline. 
However, since visual backbones are usually pre-trained on image~\cite{krizhevsky2012imagenet} and/or video datasets~\cite{kay2017kinetics} without access to text, these features are less effective for video-and-language tasks.
Besides, video feature extractors in these approaches are not finetuned on target task data, preventing features to easily adapt to different domains.
In contrast, recent methods ClipBERT~\cite{lei2021less} and FiT~\cite{Bain21} demonstrate more effective results by end-to-end finetuning the visual backbone with only a few \textit{sparsely} sampled frames.
However, ClipBERT is pre-trained with image-text data thus is less effective in aggregating information across frames, while FiT is a retrieval-specific architecture that does not naturally generalize to videoQA task.
In this regard, our \name~is the first sparse pre-training architecture that tackles both tasks while in the meantime, demonstrating the benefit of pre-training on video-text pairs.

\noindent\textbf{Video-and-language Pre-training.}~
%
%
Apart from the canonical pre-training tasks, such as masked language modeling (MLM)~\cite{devlin2018bert,sun2019videobert,zhu2020actbert,luo2020univl,li2020hero,lei2021less} and video-text matching (VTM)~\cite{li2020hero,luo2020univl},
several methods~\cite{miech2020end, luo2020univl, xu2021videoclip} apply contrastive learning on \emph{offline} extracted visual features. 
%
Without adapting the visual backbone, their ability to align cross-modal features remain limited.
%
\name~jointly learns the unimodal and multimodal encoders, thus mitigating the disconnection in-between.
In order to design effective visually-grounded pre-training tasks, VideoBERT~\cite{sun2019videobert} predicts centroids of vector quantizations of video features.
Such unsupervised quantization is noisy per se while neglecting textual cues, which curtails its capabilities in learning cross-modal interactions.
ActBERT~\cite{zhu2020actbert} uses detectors to acquire object information. In addition to their computational inefficiency, detectors pre-trained on images usually have limited categories and compromised detection results on videos.
In contrast, our proposed prompting entity modeling task is detector-free.
By exploiting the instance-level video-text alignment, we can generate reliable entity pseudo-labels with a large vocabulary,
leading to more efficient and effective learning of region-entity alignment.

\noindent\textbf{Zero-shot Visual Recognition with Prompts.}~There have been longstanding efforts to exploit text descriptions for learning visual recognition models.
These include early efforts~\cite{barnard2001clustering,lampert2009learning,farhadi2009describing} that use text to learn attributes of images;  \cite{palatucci2009zero,socher2013zero,norouzi2014zero} that map images to the pretrained text embedding space, and visual n-gram~\cite{li2017learning} that predicts text n-grams given image inputs.
%
More recently, CLIP~\cite{radford2learning} instantiates prompt templates with label text of visual categories.
It then predicts the category by computing the similarity between each image-prompt pairs.
This inspires us to the design of entity prompter.
Since the entity prompter is trained with the entire video-text corpus, during pre-training, it can provide additional entity information unavailable in the text description for each individual video-text pair,
thus leading to better entity-informed video representations.
\vspace{-0.5ex}
\section{Video-Language Pre-training with \name}
\vspace{-0.5ex}

In this section, we first introduce important constituent modules of \name~in Section~\ref{sec:method-arch}.
Then we present the pre-training objectives in Section~\ref{sec:method-pt}, with a particular focus on the proposed video-text contrastive (VTC) loss and the prompting entity modeling (PEM)
pre-training task.
We introduce pre-training datasets in Section~\ref{sec:method-ptdata}.
Lastly, we describe important implementation details in Section~\ref{sec:method-impl}.

\subsection{\textbf{\name}~Architecture}\label{sec:method-arch}
Figure 2 gives an overview of \name~'s architecture.
In particular, \name~consists of two main modules, a 
\emph{video-language pre-training model} and a \emph{prompter}.
The prompter serves to generate soft entity labels to supervise the pre-training of the video-language model.
Both modules contain their own video encoder and text encoder to extract features for video and text inputs, respectively.
The pre-training model has an additional multimodal encoder to further capture the interaction between the two modalities.
Details for each component are as follows.

\noindent\textbf{Visual Encoder.}~We use a 12-layer TimeSformer$_{224}$~\cite{timesformer} to extract video features, with $224$ the height and width of input frames. 
For $N_v$ frames sparsely sampled from each input video, TimeSformer first partitions each frame into $K$ non-overlapping patches, which are flattened and fed to a linear projection layer to produce a sequence of patch tokens.
Learnable positional embeddings are also added to the patch tokens. 
Then the TimeSformer applies self-attention along the temporal and spatial dimensions separately in order, leading to per-frame features $\Tilde{\Vec{v}}\in\mathbb{R}^{N_v\times K\times d}$, with $d$ the feature dimension.
A temporal fusion layer (\ie~mean-pooling) is applied to $\Tilde{\Vec{v}}$ along the temporal dimension to aggregate per-frame features into video features.
As the output of visual encoder, we obtain a sequence of visual embeddings: $\{\Vec{v}_\mathrm{cls}, \Vec{v}_1,...,\Vec{v}_K\}$, with $\Vec{v}_i\in\mathbb{R}^{d}$ and $\Vec{v}_\mathrm{cls}$ the embedding of the video \texttt{[CLS]} token.

\noindent\textbf{Text Encoder.}~We use a 6-layer transformer~\cite{transformer} model to represent text tokens. Given an input text description of $N_t$ tokens, the text encoder outputs an embedding sequence $\{\Vec{t}_\mathrm{cls}, \Vec{t}_1,...,\Vec{t}_{N_t}\}$, with $\Vec{t}_i\in\mathbb{R}^{d}$ and $\Vec{t}_\mathrm{cls}$ the embedding of the text \texttt{[CLS]} token. Similar to video encoder, we also add positional embeddings to the text tokens.

\noindent\textbf{Multimodal Encoder.}~We employ a 6-layer transformer to model the interaction between video and text features from the two unimodal encoders.
Since positional embeddings are already injected in each unimodal encoder, we directly concatenate video and text features to feed the multimodal transformer.
The outputs are multimodal embeddings $\{\Vec{e}_\mathrm{cls}, \Vec{e}_1,...,\Vec{e}_{N_v+N_t}\}$, with $\Vec{e}_i\in\mathbb{R}^{d}$.
For notational convenience, we drop the multimodal embedding for the video \texttt{[CLS]} token as it is not used in pre-training losses.

\subsection{Pre-training for \textbf{\name}}\label{sec:method-pt}
We pre-train \name~with four objectives, including two canonical ones, \ie~masked language modeling (MLM) and video-text matching (VTM) as in~\cite{sun2019videobert,zhu2020actbert,li2020hero,lei2021less}.
In this section, we focus on presenting the new techniques in \name, \ie~the video-text contrastive (VTC) loss and the prompting entity modeling (PEM) loss, while only briefly outlining MLM and VTM in Section~\ref{sec:method-mlm}, referring interested readers to~\cite{devlin2018bert,sun2019videobert,lei2021less} for details.

The motivation of both VTC and PEM is to strengthen cross-modal alignment between video and text. While VTC emphasizes on capturing instance-level alignment for video-text pairs, PEM encourages the model to align local video regions with textual entities.
In the following, we introduce these two pre-training objectives in order.
\vspace{-0.5em}
\subsubsection{Contrastive Video-Text Alignment}\label{sec:method-vtc}
Existing sparse video-language pre-training models use either dot-product~\cite{miech2019howto100m,miech2020end,xu2021videoclip,Bain21} or rely entirely on a transformer encoder~\cite{lei2021less,li2020hero,sun2019videobert,zhu2020actbert} to model cross-modal interactions.
However, since video and text features reside in different embedding spaces, such methods lead to less satisfactory alignment.
To this end, we present a video-text contrastive (VTC) loss to align features from the unimodal encoders before sending them into the multimodal encoder.
Specifically, given the embeddings of video and text \texttt{[CLS]} tokens, we optimize a similarity function between video $V$ and text $T$:
\begin{equation}
s(V,T)=g_v(\Vec{v}_\mathrm{cls})\cdot g_t(\Vec{t}_\mathrm{cls}),
\end{equation}
such that paired video and text descriptions have higher similarity scores, where $g_v(\cdot)$ and $g_t(\cdot)$ are linear projections that transform the \texttt{[CLS]} embeddings to a common normalized low-dimensional (\eg~256-d) space.


Following~\cite{radford2learning,align}, the contrastive loss considers matched pairs as positive and all others pairs that can be formed in a batch as negatives.
For each input video-text pair $\langle V_i,\,T_i\rangle$, the video-text contrastive loss consists of two symmetric terms, one for video-to-text classification:
\begin{equation}
    \mathcal{L}_{\mathrm{v2t}}= -\mathrm{log}\frac{\exp (s(V_i,T_i) / \tau)}{\sum_{j=1}^B \exp (s(V_i,T_j)/ \tau)} 
\end{equation}
and the other for text-to-video classification:
\begin{equation}
     \mathcal{L}_{\mathrm{t2v}}= -\mathrm{log}\frac{\exp (s(T_i,V_i)/ \tau)}{\sum_{j=1}^B \exp (s(T_i,V_j)/ \tau)},
\end{equation}
where $\tau$ is a learnable temperature parameter, and $B$ is the batch size. The video-text contrastive loss is then defined as $\mathcal{L}_{\mathrm{vtc}}=\frac{1}{2} (\mathcal{L}_{\mathrm{v2t}}+\mathcal{L}_{\mathrm{t2v}})$.

%

\subsubsection{Prompting Entity Modeling}\label{sec:method-obj}
While masked language modeling has demonstrated its effectiveness on learning token-level text representations~\cite{devlin2018bert,liu2019roberta}, it remains a challenge to design its visually-grounded counterpart.
As a result, the limited capabilities in visual reasoning adversely impact previous work on downstream tasks, especially those requiring region-level visual information such as objects.
This is in particular an issue for existing video-language pre-training models~\cite{miech2019howto100m,miech2019howto100m,miech2020end,sun2019videobert,li2020hero}, which usually retain only coarse-grained spatial information after pooling thus losing fine-grained visual cues.
One exception is ActBERT~\cite{zhu2020actbert} that attempts to use off-the-shelf object detectors to obtain regional features.
Apart from its inefficiency, detectors trained with images tend to produce compromised detection results on video inputs.
In addition, detectors are usually trained with restricted object categories (\eg~less than a hundred~\cite{lin2014microsoft}), given its prohibitive expense to scale up the laborious annotations.

We introduce \emph{prompting entity modeling} (PEM), a new visually-grounded pre-training task that improves the models' capabilities in capturing local regional information and strengthening cross-modal alignment between video regions and textual entities.
Specifically, PEM requires a \emph{prompter} module that generates soft pseudo-labels identifying entities that appear in random video crops.
The pre-training model is then asked to predict the entity categories in the video crop, given the pseudo-label as the target.

The prompter serves to produce pseudo-labels of entity categories given a video crop,
without dense annotations other than webly-sourced video-text pairs with possibly noisy alignment.
To this end, we are inspired by CLIP~\cite{radford2learning} that learns image-text alignment from noisy pairs.
Specifically, we first pre-train the prompter, which consists of two unimodal encoders, on video-text pairs with the VTC loss as in Section~\ref{sec:method-vtc}, and freeze its parameters thereafter.

The prompter maintains a predetermined list of $M$ text prompts.
Each text prompt is an instantiation of a template, \eg~``\texttt{A video of \{ENTITY\}}.'', where \texttt{ENTITY} is a frequent noun in the pre-training corpus, such as \texttt{dog}, \texttt{grass}, \texttt{sky}, etc.
After the prompter is pre-trained, it computes the \texttt{[CLS]} embedding for each text prompt as $\{\Vec{{t}}^1_{\mathrm{cls}},\Vec{{t}}^2_{\mathrm{cls}},...,\Vec{{t}}^M_{\mathrm{cls}}\}$.

To generate entity labels, given one video input, we first obtain a random video crop $\hat{V}$ (\eg~the same spatial region across sampled frames) and its \texttt{[CLS]} embedding $\Vec{\hat{v}}_{\mathrm{cls}}$ from the prompter's video encoder. The prompter then computes an entity pseudo-label $\Vec{q}_{\hat{V}}\in\mathbb{R}^{M}$ for the video crop as the softmax-normalized similarity between $\Vec{\hat{v}}_{\mathrm{cls}}$ and all the prompt embeddings $\{\Vec{{t}}^m_{\mathrm{cls}}\}_{m=1}^M$:
\begin{equation}
    q_{\hat{V},m}=\frac{\exp (s(\hat{V},T_m)/ \tau)}{\sum_{m=1}^M \exp (s(\hat{V},T_m)/ \tau)}
\end{equation}
During pre-training of the video-language model, we apply mean pooling on the embeddings from the multimodal encoder that correspond to the spatial location of the video crop $\hat{V}$, denoted as $\Vec{e}_{\hat{V}}\in\mathbb{R}^{d}$.
We use a classifier (\eg~MLP) to compute the softmax-normalized entity prediction $\Vec{p}_{\hat{V}}$.
The prompting entity modeling loss is then defined as the cross-entropy between $\Vec{p}_{\hat{V}}$ and $\Vec{q}_{\hat{V}}$:
\begin{equation}
\label{eqn:itc}
\mathcal{L}_\mathrm{pem}=-\sum_{m=1}^M q_{\hat{V},m}\cdot\mathrm{log}~p_{\hat{V},m}
\end{equation}
Prompting entity modeling features a diverge range of entities while requiring no extra human annotations, which yields an efficient and scalable solution to generate visually-grounded regional supervisions for cross-modal learning.
\vspace{-1em}
\subsubsection{Overall Pre-training Objectives}\label{sec:method-mlm}
\vspace{-0.5ex}
We also employ the widely-adopted masked language modeling (MLM) loss $\mathcal{L}_{\mathrm{mlm}}$ and video-text matching $\mathcal{L}_{\mathrm{vtm}}$ considering their effectiveness.
The MLM objective utilizes both video and the contextual text to predict the masked text tokens.
We randomly mask input tokens with a probability of $15\%$ and replace them with a special token $\texttt{[MASK]}$.
Video-text matching is a binary classification task which predicts whether a video and a text description are matched with each other.
We use the multimodal \texttt{[CLS]} token $\Vec{e}_\mathrm{cls}$ as the joint representation of the video-text pair, and trains the model with a cross entropy loss.
Negative samples are generated from non-parallel video-text pairs from the batch.
Following~\cite{ALBEF}, we employ contrastive hard negative mining to find more informative in-batch negatives for VTM.
The overall pre-training objective of \name~is:
\begin{equation}
    \mathcal{L}=\mathcal{L_\mathrm{vtc}} + \mathcal{L}_\mathrm{pem} + \mathcal{L}_\mathrm{mlm} + \mathcal{L}_\mathrm{vtm}
\end{equation}
\subsection{Pre-training Datasets}\label{sec:method-ptdata}
We pre-train our model with the webly-sourced dataset WebVid-2M~\cite{Bain21}, which contains 2.5M video-text pairs.
In addition, as suggested by ClipBERT~\cite{lei2021less} and FiT~\cite{Bain21}, pre-training with image-pairs can improve spatial representations of videos, we therefore include CC-3M~\cite{2018cc3m} into our pre-training corpus.
During pre-training, we duplicate images from CC-3M to make static videos.
This in total amounts to 5.5M video-text pairs, which is an order of magnitude less than the commonly-adopted HowTo100M~\cite{miech2020end,li2020hero,zhu2020actbert} and of a comparable size to those used in~\cite{lei2021less,Bain21}.

\subsection{Implementation Details}\label{sec:method-impl}~We implement \name~in PyTorch~\cite{paszke2019pytorch}.
In detail, we initialize both the spatial and temporal attention blocks of TimeSformer by reusing ViT-B/16 weights pre-trained on ImageNet-21k~\cite{dosovitskiy2020image}.
Text encoders are initialized using the first 6-layer of the BERT$_{\mathrm{base}}$ model~\cite{devlin2018bert}, and the multimodal encoder is initialized using the last 6-layers weights of BERT$_{\mathrm{base}}$.
We pre-train \name~for 100k iterations, roughly equivalent to 10 epochs, using a batch size of 256 on 16 NVIDIA A100 GPUs.
We use AdamW~\cite{adamw2018} optimizer with a weight decay of 0.001. The learning rate is first warmed-up to $1e^{-4}$, then it follows a linear decay schedule.
Since videos are usually of different aspect ratios, we first rescale them to $224\times 224$.
For each video, we sample 4 frames randomly as inputs to the visual encoder while preserving their orderings in-between.
For PEM, we use POS tagger\footnote{https://github.com/explosion/spaCy} and retain the top 1k most frequent nouns as the entity names.
We obtain random video crops occupying $30\%-50\%$ of the original spatial area as inputs to the prompter. We discard a pseudo-label if the most likely entity has a normalized similarity score smaller than 0.2.

\begin{table*}[htb]
    \small
    \aboverulesep = 0.55mm
    \belowrulesep = 0.55mm
	\centering	
	\resizebox{0.95\textwidth}{!}
{%
	\begin{tabular}	{l   c@{\hspace{1.5\tabcolsep}} c @{\hspace{1.5\tabcolsep}} c @{\hspace{1.5\tabcolsep}} c c @{\hspace{1.5\tabcolsep}} c @{\hspace{1.5\tabcolsep}} c @{\hspace{1.5\tabcolsep}} c @{\hspace{1.5\tabcolsep}} c @{\hspace{1.5\tabcolsep}}c
	@{\hspace{1.5\tabcolsep}}c
	@{\hspace{1.5\tabcolsep}}c 
	@{\hspace{1.5\tabcolsep}}c
	@{\hspace{1.5\tabcolsep}}c
	@{\hspace{1.5\tabcolsep}}c@{\hspace{1.5\tabcolsep}}}
		\toprule \multirow{2}{*}{\textbf{Pre-training tasks}} & \multicolumn{4}{c}{\footnotesize\textbf{MSRVTT Retrieval}}& \multicolumn{4}{c}{\footnotesize\textbf{DiDeMo Retrieval}} & \footnotesize\textbf{MSVD-QA} & \footnotesize\textbf{MSRVTT-QA} \\
	 &  \footnotesize\textbf{R1}$\uparrow$ &\footnotesize\textbf{ R5}$\uparrow$
	 &  \footnotesize\textbf{R10}$\uparrow$ &\footnotesize\textbf{ MdR}$\downarrow$
	 & \footnotesize \textbf{R1}$\uparrow$ & \footnotesize\textbf{R5}$\uparrow$
	 & \footnotesize\textbf{R10}$\uparrow$ & \footnotesize\textbf{MdR}$\downarrow$ & \footnotesize\textbf{Acc.}$\uparrow$ &  \footnotesize\textbf{Acc.}$\uparrow$\\
	 	 \midrule
      w/o pre-training & 16.5 & 42.8 & 57.9 & 7 & 9.5 & 29.1 & 42.5 & 14 & 41.5 & 39.6  \\
      MLM + VTM & 28.5 & 53.0 & 66.8 & 5 & 29.8 & 57.7 & 69.7 & 4 & 43.3 & 40.9 \\
      MLM + VTM + PEM & 30.3 & 56.7 & 67.8 & 4 & 31.0 & 61.8 & 73.5 & 3 &  \textbf{46.3} & 41.8 \\
      MLM + VTM + VTC & 32.8 & 59.2 & 70.3 & 3 & \textbf{36.8} & 64.7 & 77.4 & \textbf{2} & 45.5 & 41.9 \\
      MLM + VTM + PEM + VTC & \textbf{33.9} & \textbf{60.7} & \textbf{73.2} & \textbf{3} & 35.9 & \textbf{67.5} & \textbf{78.8} & 3 & 45.9  & \textbf{42.1} \\ 

		\bottomrule
	\end{tabular}}
    \vspace{-5pt}
	\caption
	{Evaluations of the proposed pre-training objectives on four downstream datasets. MLM: masked language modeling loss. VTM: video-text matching loss. PEM: prompting entity modeling loss. VTC: video-text contrastive loss. 
	R@k denotes recall (\%) with k retrieval efforts; MdR denotes median ranking for retrieved videos. We use acc. to denote accuracy.
	}
	\label{tbl:main}

\end{table*}		
\begin{figure*}[ht]
\centering
  \includegraphics[width=\textwidth]{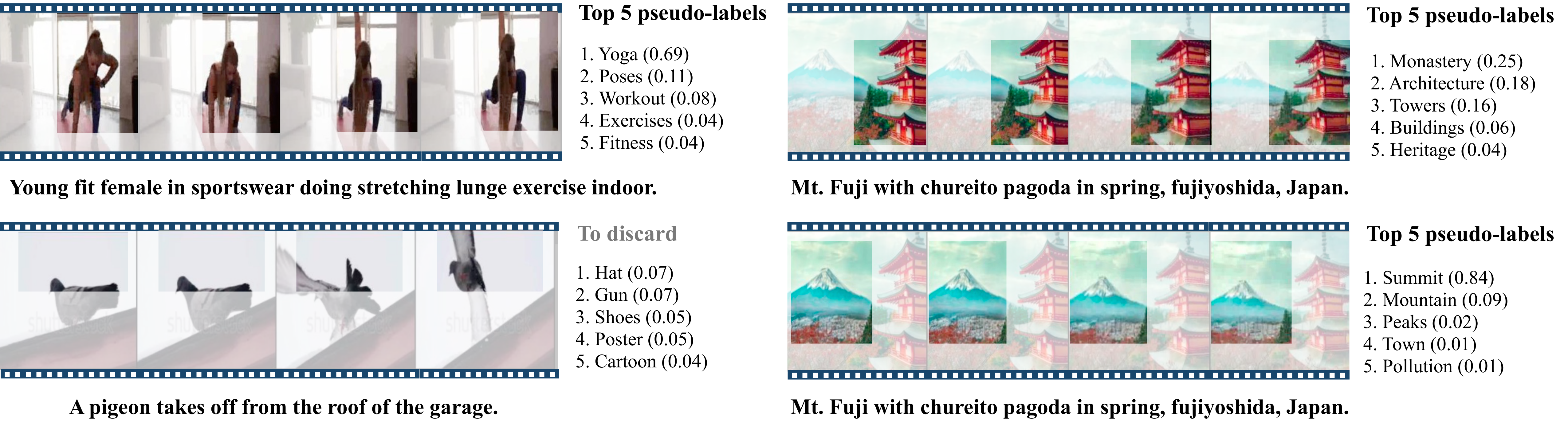}
\vspace{-15pt}
\caption{Examples of the pseudo-labels generated by the prompter (scores in bracket). The highlighted areas are fed to the prompter. Our method generates a diverse range of common entity categories that are not usually covered by object detectors, \eg~towers, summit, yoga.
Besides, entity labels do not always appear in the text description, serving as a source of corpus-level supervision.
\textbf{Bottom left}: a random crop that does not contain entities. The prompter thereby produces pseudo-labels with low similarities. During pre-training, we discard a pseudo-label if its most likely entity has a score less than 0.2. \textbf{Right}: labels generated for different crops from the same video.}
\label{fig:plabel}
\vspace{-3.5ex}

\end{figure*}

\vspace{-0.5ex}
\section{Experiments}
\vspace{-0.5ex}
We evaluate the performance of \name~on text-video retrieval and video question answering tasks across four commonly-used datasets, introduced in Section~\ref{sec:exp-setup}.
The purpose of the evaluation is three-fold. First, we demonstrate the effectiveness of major technical contributions (\ie~video-text contrastive loss and prompting entity modeling) in Section~\ref{sec:exp-method}.
We then compare the performance of \name~with previous methods, including task-specific and pre-training architectures, in Section~\ref{sec:exp-ret} and Section~\ref{sec:exp-qa}, on retrieval and question answering tasks, respectively.
Finally, we show ablation results on design choices and analyze model behaviors in Section~\ref{sec:exp-ablate}.

\subsection{Downstream Tasks and Datasets}\label{sec:exp-setup}
\noindent{\textbf{Text-Video Retrieval.}} (i) \textbf{MSRVTT}~\cite{xu2016msr} contains 10K videos with 200K text captions. We follow the common protocol~\cite{yu2018joint,miech2019howto100m,li2020hero,zhu2020actbert,lei2021less} and use 7k videos for training and report results on the 1k test split~\cite{yu2018joint}. (ii) \textbf{DiDeMo}~\cite{anne2017localizing} contains 10k videos from Flickr with 40k text descriptions. We follow~\cite{liuuse,lei2021less,luo2020univl} and evaluate paragraph-to-video retrieval, where sentence descriptions for each video are concatenated together as a single text query.
We do not use the ground-truth proposals for temporal localization to ensure fair comparisons with previous work.

\noindent{\textbf{Video Question Answering.}} We focus on the task of open-ended video question answering.
(i)~\textbf{MSVD-QA}~\cite{xu2017video} is built upon videos and text descriptions from MSVD~\cite{chen2011collecting}. The MSVD-QA dataset has in total 1,970 videos and 50k question answer pairs, with 2,423 answer candidates.
(ii)~\textbf{MSRVTT-QA}~\cite{xu2017video} is built upon videos and captions from MSRVTT, which contains 10k videos with 243k open-ended questions and 1.5k answer candidates. 

\noindent\textbf{Finetuning Setups.}~On downstream tasks, \name~allows end-to-end finetuning of the video backbone with raw video frames as input. During finetuning, we randomly sample $N$ frames per video, where $N=8$ for retrieval and $N=16$ for QA, with more ablations present in Section~\ref{sec:exp-qa}.
Temporal position embeddings in TimeSformer are interpolated to accommodate different number of input frames.
During inference, we sample frames uniformly to ensure reproducibility.
To keep pre-training and finetuning setups consistent,
we resize all the videos to $224\times 224$ before feeding them into the model. Although this does not maintain the original aspect ratios, we observe no significant performance drop as our pre-training dataset contains videos of various aspect ratios.
For finetuning on retrieval, we reuse the video-text matching head during pre-training and optimize the sum of both VTC and VTM losses.
We obtain similarity scores from the output of VTM head during inference.
For QA task, we add a simple MLP on the multimodal \texttt{[CLS]} token for classification and optmize the conventional cross-entropy loss between predictions and ground-truth answer labels.
During inference, predictions are obtained as the answer with the highest probability.
All the finetuning experiments are performed on 8 NVIDIA A100 GPUs, taking one to five hours to complete depending on the datasets.
More training details can be found in the appendix.

\vspace{-1ex}
\subsection{Evaluation on the Proposed Methods}\label{sec:exp-method}
\vspace{-0.5ex}
We first evaluate the impact of our main technical contributions (\ie~video-text contrastive loss and prompting entity modeling) in Table~\ref{tbl:main}.
Compared with pre-training using only MLM and VTM, both PEM and VTC substantially improve the performance across all the datasets.
VTC is in particular useful for the retrieval task. The reason is that the VTC loss explicitly maximizes the instance-level similarity between positive video-text pairs, which is well-aligned with the goal of retrieval.
We notice that PEM significantly improves the performance of videoQA, especially on MSVD-QA,
due to its ability to learn finer-grained regional features.
%
%
While enabling both PEM and VTC losses has complementary effects for most datasets, we also observe it leads to slightly worse accuracy on MSVD-QA.
Our observation is that MSVD-QA contains more questions requiring region-level knowledge, including object categories (\eg~dough, swords), animal species (\eg~hare,\,eagle) and scenes (\eg~river,\,cliff), which can be well modeled using PEM, rendering the impact of VTC negligible. In contrast, MSRVTT-QA involves more coarse-grained visual information such as activities.
As a result, using both PEM and VTC complements with each other on MSRVTT-QA.

\noindent\textbf{Example Pseudo-labels.} In Figure~\ref{fig:plabel}, we show examples of pseudo-labels generated by the prompter module. Our approach generates a more diverse range of entity categories beyond typical object classes from detection annotations. This is in particular beneficial when downstream tasks require a large vocabulary, such as open-ended videoQA.

\vspace{-1ex}
\subsection{Evaluation on Video-Text Retrieval}\label{sec:exp-ret}
\vspace{-0.5ex}
\begin{table}[!t]
\resizebox{0.48\textwidth}{!}{
    \aboverulesep = 0.55mm
    \belowrulesep = 0.55mm
	\centering	
     \setlength\tabcolsep{4pt}
		\begin{tabular}	{l  c |  c c c c}
			\toprule
			\textbf{Method} & \textbf{PT datasets} & \textbf{R1}$\uparrow$     & \textbf{R5}$\uparrow$ & \textbf{R10}$\uparrow$ & \textbf{MdR}$\downarrow$\\
			\midrule
		\multicolumn{6}{l}{\textbf{Finetuning}}\\
			\midrule
			JSFusion~\cite{yu2018joint} &- & 10.2 & 31.2 & 43.2 & 13 \\	HT100M~\cite{miech2019howto100m} & HT (100M) & 14.9 & 40.2 & 52.8 & 9\\	ActBERT~\cite{zhu2020actbert} & HT (100M) & 16.3 & 42.8 & 56.9 & 10\\
		NoiseEst.~\cite{amrani2021noise} & HT (100M) & 17.4 & 41.6 & 53.6 & 8\\
		HERO~\cite{fan2019heterogeneous} & HT (100M) & 16.8 & 43.4 & 57.7 & - \\
		ClipBERT~\cite{lei2021less} & \footnotesize\makecell{COCO +\\ VG (5.6M)} & 22.0 & 46.8 & 59.9 & 6\\
		AVLNet~\cite{le2020hierarchical} & HT (100M) & 27.1 & 55.6 & 66.6 & 4\\
		VideoClip~\cite{xu2021videoclip} & HT (100M) & 30.9 & 55.4 & 66.8 & -\\
		\textcolor{lightgray}{SupportSet~\cite{patrick2021supportset}} & \textcolor{lightgray}{HT (100M)} & \textcolor{lightgray}{30.1} & \textcolor{lightgray}{58.5} & \textcolor{lightgray}{69.3} & \textcolor{lightgray}{3}\\
		\textcolor{lightgray}{FiT}~\cite{Bain21} & \textcolor{lightgray}{\footnotesize{\makecell{Web2M + \\CC3M (\textbf{5.5M})}}} & \textcolor{lightgray}{31.0} & \textcolor{lightgray}{59.5} & \textcolor{lightgray}{70.5} & \textcolor{lightgray}{3} \\			\midrule
		{\textbf{\name}} & {\footnotesize{\makecell{Web2M + \\CC3M (\textbf{5.5M})}}} & \textbf{33.9} & \textbf{60.7} & \textbf{73.2} & \textbf{3} \\
		\bottomrule\toprule
		\multicolumn{6}{l}{\textbf{Zero-shot}}\\
		\midrule
		HT100M~\cite{miech2019howto100m} & HT (100M) & 7.5 & 21.2 & 29.6 & 38\\ ActBERT~\cite{zhu2020actbert} & HT (100M) & 8.6 & 23.4 & 33.1 & 36\\
		SupportSet~\cite{patrick2021supportset} & HT (100M) & 8.7 & 23.0 & 31.1 & 31\\
         MIL-NCE~\cite{miech2020end} & HT (100M) & 9.9 & 24.0 & 32.4 & 29.5\\ VideoCLIP~\cite{xu2021videoclip} & HT (100M) & 10.4 & 22.2 & 30.0 & -\\
            FiT~\cite{Bain21} & {\footnotesize{\makecell{Web2M + \\CC3M (\textbf{5.5M})}}} & {18.7} & {39.5} & {51.6} & {10} \\			\midrule
		{\textbf{\name}} & {\footnotesize{\makecell{Web2M + \\CC3M (\textbf{5.5M})}}} & {\textbf{24.1}} & {\textbf{44.7}} & {\textbf{55.4}} & 
		\textbf{8}\\
			\bottomrule
		\end{tabular}}
    \vspace{-5pt}
    \caption
	{Comparisons with existing text-to-video retrieval methods with \textbf{finetuning} and \textbf{zero-shot} setups on \textbf{MSRVTT}. We follow the common partition with 7k training videos. \emph{Methods using 9k training videos are greyed out}. Both partition protocols share the same 1k testing videos. R@k denotes recall (\%) with k retrieval efforts; MdR denotes median ranking for retrieved videos. The pre-training datasets are HowTo100M (HT)~\cite{miech2019howto100m}, MS-COCO (COCO)~\cite{lin2014microsoft}, Visual Genome (VG)~\cite{visualgenome2017}, WebVid2M (Web2M)~\cite{Bain21} and Conceptual Captions (CC3M)~\cite{2018cc3m}.}
	\label{tbl:msrvtt-ft}
\vspace{-3.5ex}
\end{table}		
\begin{table}[!t]
\resizebox{0.48\textwidth}{!}{
    \aboverulesep = 0.55mm
    \belowrulesep = 0.55mm
	\centering	
     \setlength\tabcolsep{4pt}
		\begin{tabular}	{l  c |  c c c c}
			\toprule
			\textbf{Method} & \textbf{PT datasets} & \textbf{R1}$\uparrow$     & \textbf{R5}$\uparrow$ & \textbf{R10}$\uparrow$ & \textbf{MdR}$\downarrow$\\
			\midrule	\multicolumn{6}{l}{\textbf{Finetuning}}\\
			\midrule
S2VT~\cite{venugopalan2015translating} &- & 11.9 & 33.6 & - & 13 \\	FSE~\cite{zhang2018cross} & -  & 13.9 & 36.0 & - & 11\\
		CE~\cite{liuuse} & - & 16.1 & 41.1 & - & 8\\
		MoEE~\cite{miech2018learning} & - & 16.1 & 41.2 & 55.2 & 8\\	ClipBERT~\cite{lei2021less} & \footnotesize{\makecell{COCO + \\VG (5.6M)}} & 20.4 & 48.0 & 60.8 & 6\\
		TT-CE~\cite{croitoru2021teachtext} & - & 21.6 & 48.6 & 62.9 & 6\\
		FiT~\cite{Bain21} & \footnotesize{{\makecell{Web2M + \\CC3M (\textbf{5.5M})}}} & 31.0 & 59.8 & 72.4 & 3\\
\midrule
		{\textbf{\name}} & \footnotesize{\makecell{Web2M + \\CC3M (\textbf{5.5M})}} & {\textbf{35.9}} & {\textbf{67.5}} & \textbf{78.8} & \textbf{3}\\
			\bottomrule\toprule
		\multicolumn{6}{l}{\textbf{Zero-shot}}\\
		\midrule
		VideoCLIP~\cite{xu2021videoclip} &HT (100M)& 16.6 & 46.9 & - & -\\
            FiT~\cite{Bain21} & {{\footnotesize{\makecell{Web2M + \\CC3M (\textbf{5.5M})}}}} & {21.1} & {46.0} & {56.2} & {7} \\
            \midrule
		{\textbf{\name}} &  {\footnotesize{\makecell{Web2M + \\CC3M (\textbf{5.5M})}}} & \textbf{23.8} & \textbf{47.3} & \textbf{57.9} & \textbf{6}\\
			\bottomrule
		\end{tabular}}
    \vspace{-2pt}
    \caption
	{Comparisons with existing text-to-video retrieval methods with \textbf{finetuning} and \textbf{zero-shot} setups on \textbf{DiDeMo}.
	R@k denotes recall (\%) with k retrieval efforts; MdR denotes median ranking for retrieved videos.
	}
	\label{tbl:didemo-ft}
\vspace{-1ex}
\end{table}		
\begin{table}[!t]
\resizebox{0.48\textwidth}{!}{
    \aboverulesep = 0.55mm
    \belowrulesep = 0.55mm
	\centering	
		\begin{tabular}	{l  c |  c c }
			\toprule
			\textbf{Method} & \textbf{PT datasets} & \textbf{MSRVTT}     & \textbf{MSVD} \\
			\midrule
	E-SA~\cite{xu2017video} & - & 29.3 & 27.6 \\
			ST-TP~\cite{jang2017tgif} & - & 30.9 & 31.3 \\	AMU~\cite{xu2017video} & - & 32.5 & 32.0 \\
		Co-mem~\cite{gao2018motion} & -& 32.0 & 31.7 \\
		HME~\cite{fan2019heterogeneous} & - & 33.0 & 33.7 \\
		LAGCN~\cite{huang2020location} & - & - & 34.3 \\
		HGA~\cite{jiang2020reasoning} & - & 35.5 & 34.7 \\	QUEST~\cite{jiang2020divide} & - & 34.6 & 36.1 \\	HCRN~\cite{le2020hierarchical} & - & 35.6 & 36.1 \\
		ClipBERT~\cite{lei2021less} & \footnotesize\makecell{COCO +\\ VG (5.6M)} & 37.4 & - \\
		SSML~\cite{amrani2021noise} & HT (100M) & 35.1 & 35.1 \\
	CoMVT~\cite{seo2021look} & HT (100M) & 39.5 & 42.6 \\
		\textcolor{lightgray}{VQA-T}~\cite{yang2021just} & \textcolor{lightgray}{\footnotesize{HTVQA (69M)}} & \textcolor{lightgray}{41.5} & \textcolor{lightgray}{46.3}\\			\midrule
		{\textbf{\name}} & {\footnotesize\makecell{Web2M + \\CC3M (\textbf{5.5M})}} & \textbf{42.1} & \textbf{45.9} \\
			\bottomrule
		\end{tabular}}
    \vspace{-5pt}
    \caption
	{Comparisons with existing methods on \textbf{MSRVTT-QA} and \textbf{MSVD-QA} in top-1 accuracy (\%). VQA-T~\cite{yang2021just} uses 69M QA domain-specific data to pre-train their model while \name~uses an order of magnitude less video-text pairs from the web.
	}
	\label{tbl:qa}
\vspace{-10pt}
\end{table}
In Table~\ref{tbl:msrvtt-ft} and Table~\ref{tbl:didemo-ft}, we compare \name~with existing methods using finetuning and zero-shot text-to-video retrieval on MSRVTT and DiDeMo datasets, respectively.
\name~surpasses previous methods by a significant margin while exploiting orders of magnitude less video-text pairs with no human-written texts required.
On both datasets, \name~obtains more than $6\%$ lift in terms of R10 scores.
Note that we do not include comparison with the recent paper~\cite{luo2021clip4clip} which utilizes the powerful encoders from CLIP~\cite{radford2learning} (pretrained on 400M image-text pairs) and further post-trains them on HowTo100M~\cite{miech2019howto100m}.
Since~\cite{radford2learning} uses the same objective as VideoClip~\cite{xu2021videoclip},
the improvement over VideoClip validates the advantage of \name~pre-training.
%

\vspace{-1ex}
\subsection{Evaluation on Video Question Answering}\label{sec:exp-qa}
\vspace{-0.5ex}
Table~\ref{tbl:qa} compares \name~with existing methods on open-ended video question answering datasets MSRVTT-QA and MSVD-QA.
Most competitors have QA-specific architectures while that of \name~is generic for other video-language tasks, such as retrieval.
We obtain on-par results with VQA-T~\cite{yang2021just}, which exploits 69M \emph{QA-specific} domain data for pre-training.
In contrast, \name~uses only 5.5M video-text pairs from the web without domain knowledge.
\name~surpasses other methods by a substantial margin, with $2.6\%$ and $3.3\%$ lift in accuracy.
This demonstrates the competitive visual reasoning ability of \name.

\vspace{-0.5ex}
\subsection{Ablations and Analysis}\label{sec:exp-ablate}
\vspace{-0.5ex}
\noindent\textbf{Prompt design and ensembling.}~Similar to~\cite{radford2learning}, we observe that it is important to design and ensemble prompts with multiple templates.
Without much engineering effort, we employ a preliminary set of prompt templates, such as ``\texttt{A video of a \{ENTITY\}}'',  ``\texttt{A footage of one \{ENTITY\}}'' for video inputs; ``\texttt{A photo of a \{ENTITY\}}'' and ``\texttt{A picture of the \{ENTITY\}}'' for image inputs.
In total, we design 12 templates for video and image inputs each.
We build the ensemble by averaging over the $\Vec{t}_{\mathrm{cls}}$ embeddings of prompts instantiated with the same entity.
The effect of prompt ensembling is shown in Table~\ref{tbl:prompt-ens}.
Despite our minimal engineering efforts (we only experimented with a single set of templates), prompt ensembling demonstrates its importance in generating high-quality pseudo-labels.
It is our future work to explore more prompt engineering  strategies.
%
\begin{table}[!t]
\resizebox{0.48\textwidth}{!}{
    \aboverulesep = 0.55mm
    \belowrulesep = 0.55mm
	\centering	
     \setlength\tabcolsep{4pt}
		\begin{tabular}	{l| cc c  c c c c}
			\toprule
			\multirow{2}{*}{\textbf{}} & \multicolumn{3}{c}{\textbf{MSRVTT-FT}} &\multicolumn{3}{c}{\textbf{MSRVTT-ZS}} & {\textbf{MSVD-QA}} \\&	\textbf{R1}$\uparrow$ &\textbf{R10}$\uparrow$ & \textbf{MdR}$\downarrow$ & \textbf{R1}$\uparrow$     & \textbf{R10}$\uparrow$ & \textbf{MdR}$\downarrow$ &  \textbf{Acc.}$\uparrow$\\
			\midrule	w/o ens. & 32.7 & 73.1 & 3 & 22.6 & 52.3 & 9 &  45.0\\	
			with ens. & \textbf{33.9} & \textbf{73.2} & \textbf{3} & \textbf{24.1} & \textbf{55.4} & \textbf{8} & \textbf{45.9} \\
			\bottomrule
		\end{tabular}}
    \vspace{-4pt}
    \caption
	{Effect of \name~pre-training with and without prompt ensembling (ens.) on \textbf{MSVD-QA} and \textbf{MSRVTT text-video retrieval} with finetuning (FT) and zero-shot (ZS) setups.}
	\label{tbl:prompt-ens}
\vspace{-0.5ex}
\end{table}		

\noindent\textbf{Effect of number of entities.}
\begin{table}[!t]
\resizebox{0.48\textwidth}{!}{
    \aboverulesep = 0.55mm
    \belowrulesep = 0.55mm
	\centering	
     \setlength\tabcolsep{4pt}
		\begin{tabular}	{c| cc c  c c c c}
			\toprule
			\multirow{2}{*}{{\#ent.}} & \multicolumn{3}{c}{\textbf{MSRVTT-FT}} &\multicolumn{3}{c}{\textbf{MSRVTT-ZS}} & {\textbf{MSVD-QA}} \\&	\textbf{R1}$\uparrow$ &\textbf{R10}$\uparrow$ & \textbf{MdR}$\downarrow$ & \textbf{R1}$\uparrow$     & \textbf{R10}$\uparrow$ & \textbf{MdR}$\downarrow$ &  \textbf{Acc.}$\uparrow$\\
				\midrule $\emptyset$ & 32.8 & 70.3 & 3 & 22.6 & 53.0 & 9 & 45.5
			\\
			500 & 33.0 & 71.9 & 3 & 22.7 & 54.1 & {8} & 45.6 \\
			1000 & 33.9 & {73.2} & 3 & {24.1} & {55.4} & {8} & 45.9 \\
			2000 &  34.7 & 72.4 & 3 & 22.4 & 52.8  & 9 & 45.3 \\
			\bottomrule
		\end{tabular}}
    \vspace{-4pt}
    \caption
	{Effect of the number of entities for PEM. We report results on \textbf{MSVD-QA and MSRVTT text-video retrieval} with  finetuning (FT) and zero-shot (ZS) setups. The first row corresponds to the model trained with MLM+VTM+VTC (i.e w/o PEM).}
	\label{tbl:ablate-num-ent}
\vspace{-1ex}
\end{table}		
\begin{table}[!t]
\resizebox{0.48\textwidth}{!}{
    \aboverulesep = 0.55mm
    \belowrulesep = 0.55mm
	\centering	
     \setlength\tabcolsep{4pt}
		\begin{tabular}	{c| cc c  c c c c}
			\toprule
			\multirow{2}{*}{{\#frms}} & \multicolumn{3}{c}{\textbf{MSRVTT-FT}} &\multicolumn{3}{c}{\textbf{MSRVTT-ZS}} & {\textbf{MSVD-QA}} \\&	\textbf{R1}$\uparrow$ &\textbf{R10}$\uparrow$ & \textbf{MdR}$\downarrow$ & \textbf{R1}$\uparrow$     & \textbf{R10}$\uparrow$ & \textbf{MdR}$\downarrow$ &  \textbf{Acc.}$\uparrow$\\
			\midrule 2 & 25.7 & 63.9 & 5 & 17.3 & 48.9 & 11& 43.8\\	
			4 &31.0 & 69.6 & 4 & 21.4 & 54.4 & 8 & 44.5 \\
			8 & 33.9 & \textbf{73.2} & \textbf{3} & 24.1 & \textbf{55.4} & 8 & 45.4 \\
			16 &  \textbf{34.2} & 72.6 & \textbf{3} &\textbf{24.7} & 55.0 & \textbf{7} & 45.9\\
			\bottomrule
		\end{tabular}}
    \vspace{-5pt}
    \caption
	{Effect of the number of frames on \textbf{MSRVTT text-video retrieval and MSVD-QA}. More frames generally lead to better performance with 8-16 frames achieve a good trade-off between metrics and computation expense.}
	\label{tbl:ablate-num-frms}
\vspace{-3.5ex}
\end{table}		
We investigate the effect of the number of entities for PEM in Table~\ref{tbl:ablate-num-ent}. Compared with the model pre-trained with MLM+VTM+VTC, adding PEM brings consistent improvement with frequent entities.
This suggests that the underlying principle of PEM to learn better region-entity alignment plays the essential role in its effectiveness.
However, adding more low-frequency entities introduces noises in generating entity pseudo-labels, thus harming the pre-training.

\noindent\textbf{Effect of number of frames.} 
In Table~\ref{tbl:ablate-num-frms}, we show the results on downstream tasks with different numbers of input frames.
Generally more frames lead to better performance, while such benefit saturates with more than 8 frames on the retrieval task.
By sparsely sampling frames from the video and enabling end-to-end training of the visual backbone, \name~learns more effective representations than previous methods that use fixed offline features.



\vspace{-0.5ex}
\section{Conclusion}
\vspace{-0.5ex}
\label{sec:conclusion}
This paper proposes \name~, a new video-language pre-training framework that operates on sparsely-sampled video frames.
\name~introduces video-text contrastive learning to align instance-level unimodal features, and prompting entity modeling for fine-grained region-entity alignment. 
We verify the efficiency and efficacy of \name~ on multiple downstream datasets,
where \name~achieves substantial performance improvement over existing models.

We believe \name~opens up a new direction for vision-language research, by exploiting the uptrending technique of prompting to generate semantic pseudo-labels.
Here we list two potential ideas that can be further explored to improve \name:
(1) better prompt engineering / prompt tuning to improve the quality of entity pseudo-labels; 
(2) prompt-guided region selection with temporal information taken into consideration, which might improve the current way of random region selection.
Last but not least, \name~is not restricted to the video domain, and can be naturally extended to image-text representation learning, or even image representation learning.

\noindent\textbf{Limitations and Broader Impacts.}
While Section~\ref{sec:conclusion} discusses the technical aspect and proposes potential improvements,
here we highlight the potential negative societal impact.
Our pre-training data is collected from the web, which may contain unsuitable videos, harmful texts, and private information,
which could leak into the pre-trained models.
Additional model analysis is necessary before deployment.

{\small
\bibliographystyle{ieee_fullname}
\bibliography{egbib}
}

\section*{A. Implementation Details}

\subsection*{A.1. Prompt Templates}
Inspired by~\cite{radford2learning}, we ensemble multiple prompts templates to improve the quality of the generated pseudo-labels.
We create the ensemble over the embedding space. Namely, we compute the average embedding of all the prompts for each entity.
This allows scalable ensembling with more templates with the same computation cost as using a single prompt.
During pre-training, the dataloader switches between video and image inputs. And we employ the following prompt templates for videos and images.

\begin{table}[h]
\centering
\footnotesize
{
    \aboverulesep = 0.55mm
    \belowrulesep = 0.55mm
	\centering	
		\begin{tabular}	{l |l}
			\toprule
			\textbf{Prompts for video inputs} & \textbf{Prompts for image inputs}\\
			\midrule
	A footage of a \{\}. & A photo of a \{\}.\\
	A footage of the \{\}. & A photo of the \{\}.\\
	A footage of one \{\}. &A photo of one \{\}.\\
	A video of a \{\}. & A picture of a \{\}.\\
	A video of the \{\}. &A picture of the \{\}. \\
	A video of one \{\}. & A picture of one \{\}.\\
	A portrait of a \{\}. & A good photo of a \{\}.\\
	A portrait of the \{\}. & A good photo of the \{\}.\\
	A portrait of one \{\}. & A good photo of one \{\}.\\
	A video footage of a \{\}. & A good picture of a \{\}. \\
	A video footage of the \{\}. &A good picture of the \{\}.\\
	A video footage of one \{\}. &A good picture of one \{\}.\\
			\bottomrule
		\end{tabular}}
    \vspace{-5pt}
    \caption
	{Prompt templates used for video and image inputs.
	}
	\label{tbl:msvd-qa}
\vspace{-10pt}
\end{table}

\subsection*{A.2. Finetuning Setups}
In this section, we describe implementation details for fine-tuning the pre-trained model. For all the downstream datasets, we resize video frames to $224\times224$. During finetuning, we randomly select $N_v$ frames from the video, with parameters $\frac{N_v}{2}$ frames from the first and second half of the video, respectively. We use RandomAugment~\cite{cubuk2020randaugment}.
We apply the same augmentation across frames sampled from one video.
The default settings for finetuning on each dataset are in Table~\ref{tbl:msrvtt-ret}.
During inference, we do not use augmentation and sample uniformly. In all the experiments, we use the same random seed (\ie~$42$) to ensure reproducibility.
\begin{table}[!h]
\centering
\vspace{-2ex}
\footnotesize
{
    \aboverulesep = 0.55mm
    \belowrulesep = 0.55mm
	\centering	
		\begin{tabular}	{l| l | l}
			\toprule
			\textbf{config} & \textbf{MSRVTT} & \textbf{DiDeMo} \\
			\midrule
	optimizer & AdamW & AdamW \\
	base learning rate & 2.5e-5 & 4e-5 \\
	weight decay & 1e-3 & 1e-3 \\	optimizer momentum & $\beta_1$,\,$\beta_2$=0.9,\,0.98 & $\beta_1$,\,$\beta_2$=0.9,\,0.98\\
	learning rate schedule & linear decay & linear decay \\
	batch size & 64 & 96\\
	max. text length & 40& 50 \\
	frame number & 8 & 8 \\
	warmup ratio & 0.1 & 0.1\\
	training epochs & 5 & 10\\
	augmentation & RandAug(2,\,5)& RandAug(2,\,5) \\
	gradient accumulate step & 1 & 1\\
			\bottomrule
		\end{tabular}}
    \vspace{-5pt}
    \caption
	{End-to-end finetuning configurations for \textbf{MSRVTT} and \textbf{DiDeMo text-to-video retrieval}.
	}
	\label{tbl:msrvtt-ret}
\vspace{-10pt}
\end{table}


\begin{table}[h]
\centering
\footnotesize
{
    \aboverulesep = 0.55mm
    \belowrulesep = 0.55mm
	\centering	
		\begin{tabular}	{l| l | l}
			\toprule
			\textbf{config} & \textbf{MSRVTT} & \textbf{MSVD}\\
			\midrule
	optimizer & AdamW & AdamW \\
	base learning rate & 5e-5 & 5e-5\\
	weight decay & 1e-3 & 1e-3\\	optimizer momentum & $\beta_1$,\,$\beta_2$=0.9,\,0.98& $\beta_1$,\,$\beta_2$=0.9,\,0.98 \\
	learning rate schedule & linear decay & linear decay\\
	batch size & 96 & 96\\
	max. text length & 40 & 40 \\
	frame number & 16& 16 \\
	warmup ratio & 0.1 & 0.1\\
	training epochs & 10 & 15\\
	augmentation & RandAug(2,\,5)& RandAug(2,\,5) \\
	gradient accumulate step & 2 & 2\\
			\bottomrule
		\end{tabular}}
    \vspace{-5pt}
    \caption
	{End-to-end finetuning configurations for \textbf{MSRVTT-QA} and \textbf{MSVD-QA}.
	}
	\label{tbl:msrvtt-qa}
\vspace{-10pt}
\end{table}


\begin{figure*}[ht]
\centering
  \includegraphics[width=\textwidth]{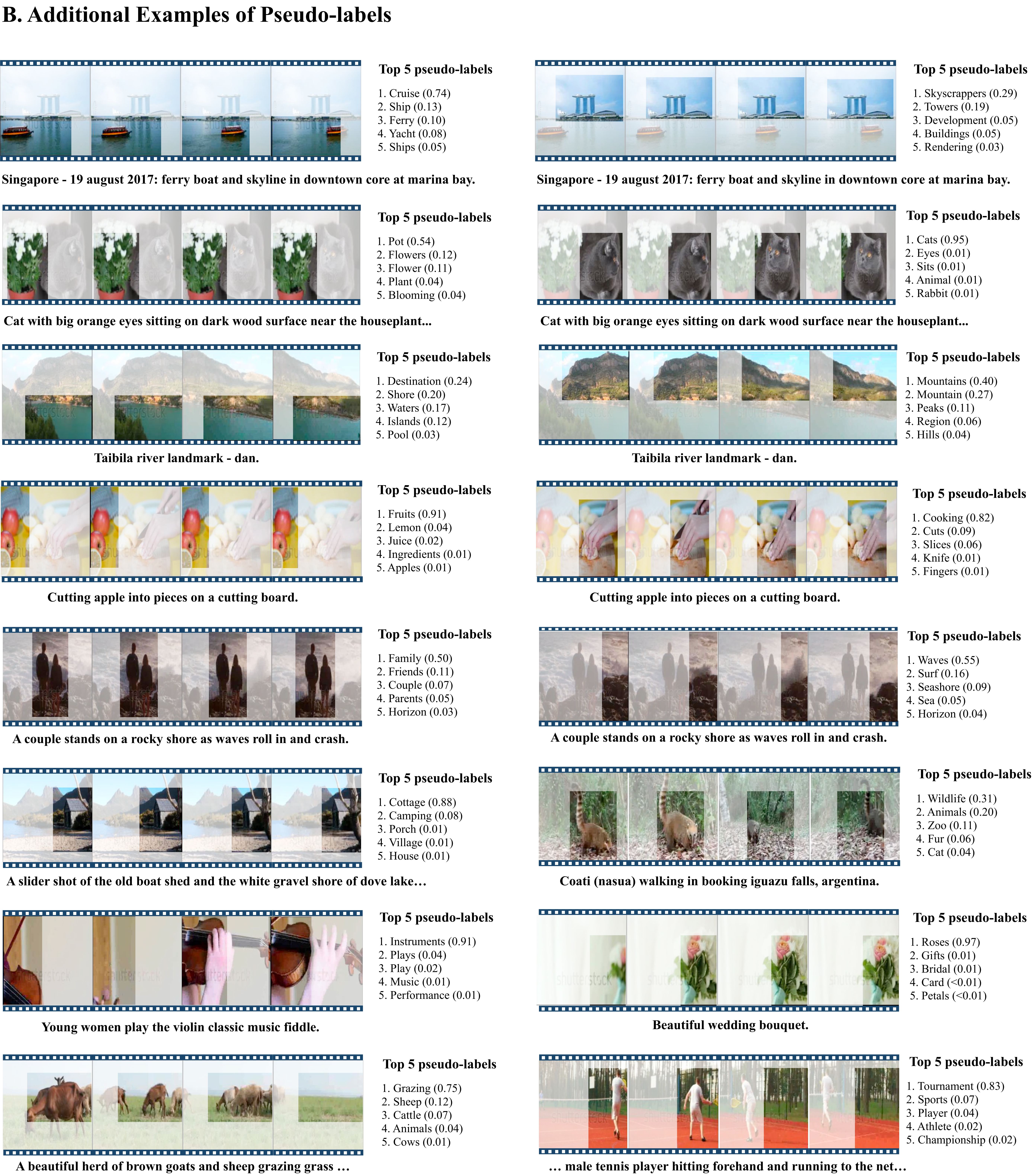}
\caption{Examples of the pseudo-labels generated by the prompter (scores in bracket). The highlighted areas are fed to the prompter. Our method generates a diverse range of common entity categories that are not usually covered by object detectors, \eg~towers, summit, yoga.
Besides, entity labels do not always appear in the text description, serving as a source of corpus-level supervision.}
\label{fig:supp-plabel}

\end{figure*}

\end{document}